\documentclass[runningheads]{llncs}

\usepackage{graphicx}
\usepackage{wrapfig}
\usepackage{url}
\usepackage{subcaption}
\usepackage[utf8]{inputenc}
\usepackage[T1]{fontenc}
\usepackage{hyperref}

\usepackage{amsmath}
\usepackage{amssymb}
\usepackage{mathptmx}

\usepackage{multicol}

\DeclareMathAlphabet{\mathcal}{OMS}{cmsy}{m}{n}

\DeclareMathOperator{\EX}{\mathbb{E}}

\begin{document}
\title{Generative Adversarial Networks: recent developments}
%



\author{Maciej Zamorski\inst{1,2} \and
Adrian Zdobylak\inst{1} \and
Maciej Zięba\inst{1,2} \and
Jerzy Świątek\inst{1} }
\institute{Wrocław University of Science and Technology, Wrocław, Poland \and
Tooploox Ltd., Wrocław, Poland}

\authorrunning{M. Zamorski et al.}

\maketitle              
\begin{abstract}
In traditional generative modeling, good data representation is very often a base for a good machine learning model. It can be linked to good representations encoding more explanatory factors that are hidden in the original data. With the invention of Generative Adversarial Networks (GANs), a subclass of generative models that are able to learn representations in an unsupervised and semi-supervised fashion, we are now able to adversarially learn good mappings from a simple prior distribution to a target data distribution. This paper presents an overview of recent developments in GANs with a focus on learning latent space representations.  

\end{abstract}
\section{Introduction}
Generative Adversarial Networks (GANs) \cite{goodfellow2014generative} are a class of generative models that can transform vectors of generated noise into synthetic samples resembling data gathered in the training set. GANs have been successfully applied to image generation \cite{radford2015unsupervised,karras2017progressive,brock2018large}, semi-supervised learning \cite{odena2016semi,zieba2017training,zamorski2018semisupervised}, domain adaptation \cite{choi2017stargan,kim2017learning,zhang2017stackgan,zhang2017stackgan++}, generation controlled by attention \cite{zhang2018self} and compression \cite{agustsson2018generative}. Currently, together with variational autoencoders (VAEs) \cite{kingma2013auto,fabius2014variational,burda2015importance,tolstikhin2017wasserstein,makhzani2015adversarial,zamorski2018adversarial}, GANs are one of the most popular and researched topic in generative modelling \cite{goodfellow2016deep,kurach2018gan,creswell2018generative}. However, correct evaluation of the GANs has been proven to be particularly difficult due to no consistent metric and inability to compute the generator probability of arbitrary samples \cite{lucic2017gans}. In this work, we provide an overview of existing GAN models, starting from the basic architectures and finishing with the complex approaches focused on particular generative tasks.

\section{Generative Adversarial Networks}
Generative Adversarial Models (GANs)\cite{goodfellow2014generative} in the last years have become a frequent choice for a task of approximating data distribution.

The basic concept of the model is taken from the game theory and assumes two competing networks, a discriminator $D$ and a generator $G$.

The role of the discriminator $D$ is to distinguish between true samples taken from data and fake samples generated by generator G.
While the network $D$ continually improves on differentiating, the generator network $G$ learns to produce better and better samples.

In practical applications, the problem is solved by updating the parameters of discriminator and generator in alternating steps.
Formally, the problem can be defined as a following min-max game:

\begin{equation}
\label{eq:value-fun-G-D-base}
\min_{G} \max_{D} V(G, D)
  = \EX_{\mathbf{x} \sim p_x}[\log D(x)]
  + \EX_{\mathbf{z} \sim p_z}[\log (1 - D(G(z)))]
\end{equation}

In early steps of learning procedures $G$ might generate poor samples. In such scenario, $D$ is expected to recognise majority of samples generated by $G$, what might lead to $log(1 - D(G(z)))$ saturation. Thus, when $1 - D(G(z))$ converges to 0 it cause generator gradient to vanish. To overcome this issue, it is advised to instead maximize $D(G(z))$, what motivates equation:

\begin{equation}
\label{eq:value-fun-G-D-final}
\max_{G} \max_{D} V(G, D)
  = \EX_{\mathbf{x} \sim p_x}[\log D(x)]
  + \EX_{\mathbf{z} \sim p_z}[\log (D(G(z)))]
\end{equation}

Despite the unquestionable potential, GANs have had few limitations, such as being unstable to train and difficult to scale. In recent years, convolutional neural networks(CNNs) have proved to be a very powerful tool for image processing. State of the art CNNs architectures consists of dozens of hidden layers, but such deep architectures did not work well with GANs. In \cite{radford2015unsupervised} authors proposed a set of good practices for Deep Convolutional GANs training. It is advised to: avoid fully connected layers in deeper architectures, use batch normalization in generator and discriminator networks, replace pooling layers with strided convolutions for discriminator and with fractional-strided convolutions for a generator. As activation function, ReLU should be used in a generator for almost all layers except last one, where $tanh(\cdot)$ is proposed. In discriminator, on the other hand, \emph{LeakyReLU} is worth considering. Described recommendations by no means should be treated as fixed rules, but instead might be a good starting point. 
Further enhancements were proposed and evaluated in \cite{salimans2016improved}. To avoid generator network mode collapse, instead of optimizing expected value with a focus on discriminator's output, authors optimize it on discriminator's intermediate layer representing hidden features. By using a feature layer, a generator is believed to generate data with respect to the distribution of real data more accurately.


In vanilla GAN, a discriminator is being trained on each example independently.
The second idea is to allow discriminator to look at multiple examples, in order to let a generator create more diverse examples.
Feature vector $f(x_i) \in \mathbb{R}^A$ of an input $x_i$ is multiplied by transformation tensor $T \in \mathbb{R}^{A \times B \times C}$ aggregating weights for similarity learning, which results in is matrix $M_i \in \mathbb{R}^{B \times C}$. Every row in $M_i$ is then compared to corresponding rows in other matrices $M_{j}$ by calculating the distance based on $L_1$ norm. This operation creates n vectors $o_i(x_i) \in \mathbb{R}^B$, which are afterwards concatenated with input $f(x_i)$ and fed to next discriminator layer.
The concept is called \textit{minibatch discrimination}.

\section{Conditional generation}
So far, no information about class or label has been regarded. The only distinction made was related to distribution the data came from. Not all problems shall be resolved by one-to-one mapping, for some of them (e.g., tagging images with keywords) more natural way is to create a one-to-many mapping.  GANs can be extended by conditioning both networks G and D on some auxiliary information $y$\cite{mirza2014conditional}:

\begin{equation}
\label{eq:value-fun-conditional}
\max_{G} \max_{D} V(G, D)
  = \EX_{\mathbf{x} \sim p_x}[\log D(x|y)]
  + \EX_{\mathbf{z} \sim p_z}[\log (D(G(z|y)))]
\end{equation}

In comparison to Conditional GAN, in AC-GAN\cite{odena2016conditional} discriminator does not utilize information about the class directly. Apart from the noise $z \sim p_z$ every sample has knowledge about the class $c \sim p_c$. For image $X$ let us denote $S$ as a source (data or generated distribution) and $C$ as a class label. The discriminator output is a modeled probability not only of $P(S|X)$ (like in vanilla GAN) but also of $P(C|X)$.

$D$ is trained to optimize $L_C + L_S$ (as presented in Equations \ref{eq:acgan_ls} and \ref{eq:acgan_lc}) and G is trained to optimize $L_C - L_S$. AC-GAN is capable of splitting dataset by classes and training G and D accordingly to subsets, as well as performing semi-supervised learning by ignoring loss component from class labels.

\begin{equation}
    \label{eq:acgan_ls}
    L_S  = \EX_{\mathbf{x} \sim p_x}[\log D(x)]
         + \EX_{\mathbf{z} \sim p_z}[\log (D(G(z)))]
\end{equation}
\begin{equation}
    \label{eq:acgan_lc}
    L_C  = \EX_{\mathbf{x, c} \sim p_{x, c}}[\log D(c|x)]
         + \EX_{\mathbf{z, c} \sim p_{z, c}}[\log (D(G(c|z)))]
\end{equation}


The improvement in numerical measures does not always go along with the improvement in human perception. The statement is particularly applicable in the case of image generation. Superresolution GAN \cite{ledig2017photo} is designed to upscale low resolution (LR) images to high-resolution (HR) by a scale factor of 4, with the utmost care for details. Authors introduce deep ResNet adapted to GAN concept and then propose using novel loss function to increase image fidelity. To create a training set, a Gaussian filter is applied to every high-resolution image $I^{HR}$, and then the image is downsampled to low-resolution image $I^{LR}$. Generator network G is trained as a supervised deep Residual Network to estimate for given LR image an HR one.
The proposed perceptual loss is defined on the activation layer of pretrained VGG19 network \cite{simonyan2015very}.
Final results are evaluated with a peak signal-to-noise ratio (PSNR) and structural similarity index (SSIM) as well as mean opinion score (MOS) - to quantify results with the help of human raters. Even though PSNR and SSIM scores are lower in every (out of three) conducted experiments, obtained MOS scores are respectively 6.2\%, 24.8\%, and 55.5\% higher, what proves that mentioned numerical metrics are not sufficient to evaluate generated images.

Conditional GANs are providing additional information (class, domain-specific image) to the generator and obtain the particular type of generated images. Moreover, there are approaches, like \emph{InfoGAN} \cite{chen2016infogan}, that aims at discovering some important latent components that have an influence on a generated image in purely unsupervised mode. Practically, it means that among space z in the generator we can distinguish some key features that have a significant influence on particular characteristics of generated objects, like shape, rotation or category. This goal is achieved by incorporating into adversarial training an additional term, that aims in increasing the mutual information between particular features delivered on the input of the discriminator and the generated image. After training the model, we are capable of controlling the generative process (shape, color rotation of the generated objects) by manipulating of the latent factors that were used to increase the mutual information.  

\section{Image to Image translation}

GAN-based models are successively applied to image-to-image translation tasks. For this particular problem, we aim at transferring some properties of the images from so-called target domain $\mathcal{X}$ to some target domain $\mathcal{Y}$ characterized by some particular features. 
Image-to-image translation models can be applied to transfer a segmented image to real good-looking scenery, can be utilized to transfer street views to the maps or used to create real images from hand-drawn sketches.

We can distinguish two approaches to the problem in terms of data availability used for training. In the first group we assume that models are trained using pairwise data, what practically means that we have access to the pairs of images, $\mathcal{D}=\{x_n,y_n\}$, from the two domains, $\mathcal{X}$ and $\mathcal{Y}$, where $x_n \in \mathcal{X}$ and $y_n \in \mathcal{Y}$. For the second group of approaches, we have only the access to unpaired sets of images from the domains, $X=\{x_n\}$ and $Y=\{y_n\}$.


One of the most promising generative models that utilizes pairwise data for image-to-image translation is \emph{Pix2Pix} \cite{isola2017image}. 
The main idea of this approach is based on conditional GAN model \cite{mirza2014conditional} where additional conditioning unit is included on the input of the generator to sample more specific objects. For this particular case generator (G) takes the example $x_n$ from domain $\mathcal{X}$ and tries to generate the corresponding image from domain $\mathcal{Y}$. The discriminator is trained to distinguish between synthetic samples generated from that domain $\mathcal{Y}$ and the corresponding samples from the domain $\mathcal{X}$ used for conditioning in generative part of training. To keep the consistency between generated and true examples in a target domain $\mathcal{Y}$, we utilize $L1$ reconstruction loss to force generated $G(x_n)$ and corresponding true samples $y_n$ to be close in data space.  

\emph{Pix2Pix} model operates on paired data from the domains. 
Here we present the architecture of CycleGAN \cite{zhu2017unpaired} that operates on unpaired images from the domains.
The structure of that model is composed of four neural networks: two domain-specific discriminators, $D_X$, and $D_Y$, and two generative networks, $G$ that generates objects from domain $\mathcal{X}$ to domain $\mathcal{Y}$, and $F$, that transfers objects from  $\mathcal{Y}$ to $\mathcal{X}$. The role of the discriminator $D_Y$ is to distinguish between true database samples from domain $\mathcal{Y}$ and those generated by model $G$. The role of the generator $G$ is to create images indistinguishable by $D_Y$. The analogical adversarial training is performed between discriminator $D_X$ and generator $F$ in the $\mathcal{X}$ domain. To obtain the cycle consistency between generated images from various domains two additional $L1$ reconstruction losses are incorporated into the training framework. The first lost is minimizing the distance between image $x$ and corresponding reconstruction $F(G(x))$. The second loss aim at minimizing distance in $\mathcal{Y}$ domain, between $G(F(y))$ and $y$. 

\section{Feature extraction via learning hidden representation}
In their original form, Generative Adversarial Networks (GANs)\cite{goodfellow2014generative} provide only a framework to generate data based on latent feature vector. A natural question comes to mind: "how can we obtain the latent representation that may be used to generate specified images?" If there is a way to produce a mapping from latent distribution to data distribution, there should be a way to perform an `inverse` operation. That would allow GANs to be used in an unsupervised manner to learn rich distributions about arbitrary data. However, the original model does not have a way to do that mapping.


For this purpose Bidirectional GANs (BiGANs)\cite{donahue2016adversarial}/Adversarially Learned Inference (ALI)\cite{dumoulin2016adversarially} were created. 
In addition to the existing Generator $G$ the BiGAN model proposes a novelty that comes from equipping the architecture with the Encoder $E$, which maps the data distribution $\mathbf{x}$ to its latent representation $\mathbf{z}$. Thus the Discriminator $D$ in BiGAN now has to discriminate not only in the data space ($\mathbf{x}$ vs $G(\mathbf{z}$) but also in the feature space ($E(\mathbf{x})$ vs $\mathbf{z}$). 

The optimization problem is now $\min_{G,E} \max_{D} V(D,E,G)$, where training objective $V(G, D, E)$ is given as:
\begin{equation}
V(G, D, E) = \EX_{\mathbf{x} \sim p_x}[\EX_{\mathbf{z} \sim p_{E}(\cdot | x)}[\log D(x, z)]] + \EX_{\mathbf{z} \sim p_z}[\EX_{\mathbf{x} \sim p_{G}(\cdot | z)}[\log (1 - D(x, z))]]
\end{equation}
 
The objective is optimized in a similar way to the original GAN approach, but with a key difference: there is no more 'real' and 'generated' data, as the Encoder $E$ and Generator $G$ now works together to fool the Discriminator $D$. However, the Encoder $E$ and the Generator $G$ do not see each other outputs. Their gradients come purely from the Discriminator decisions. However, as authors\cite{donahue2016adversarial} point out, that in order to fool the Discriminator, the Encoder and the Generator must learn to invert each other.
 
Metric learning is a task of learning the function of a distance between two given objects. It's objective is to model such mapping from data distribution $p(\mathbf{x})$ to latent distribution $p(\mathbf{z})$ that for two objects $x_1, x_2 \sim p(\mathbf{x})$, the metric returns small values for similar objects and high values for dissimilar ones. It is used in situations, where defining explicit distance function is impossible, due to a low amount, high-dimensionality or complexity of the data.

One of the first machine learning models that performed distance calculation used a type of neural networks, also called Siamese Networks \cite{bromley1994signature} or its variants \cite{chopra2005learning,hadsell2006dimensionality}. Siamese Network operated on the pair of the images and had training objective that favored small distances for objects belonging to the same group and large distance when they belong to different groups. Due to the lack of providing the context for image pair, the representations learned by the network give poor results, when used to other tasks, such as classification. 

A solution to this problem was provided in Triplet Networks \cite{hoffer2015deep}. The authors propose a simple method to add context to presented images by providing as an input to the networks three objects denoted as $x$, $x^+$ and $x^-$, where $x$ and $x^+$ were labeled as belonging to the same class while $x$ and $x^-$ were labeled as different classes. Now, denoting the features inferred from the model $T$ from the object $x$ as $T(x)$, the learning objective can be formulated as 

\begin{equation} 
L(d_+, d_-) = \lVert (d_+, d_- - 1) \rVert^2_2,
\label{eq:triplet_loss}
\end{equation} with $d_+$ and $d_-$ defined as 

\begin{equation}
d_\pm = \frac{\exp( \lVert T(x) - T(x^\pm)  \rVert_2)}{\exp( \lVert T(x) - T(x^+)  \rVert_2) + \exp( \lVert T(x) - T(x^-)  \rVert_2)}
\end{equation}

Authors use the model to solve the task of approximating data similarity \cite{chechik2010large}. For this purpose, the data already comes in the form of triplets that describe semantic closeness of samples in the dataset. Also, by giving the context, the triplet network can accurately compose a metric that is able to infer representation to be used in classification and retrieval tasks. 

In \cite{salimans2016improved,odena2016semi} authors present an approach to train GANs in a semi-supervised manner due to a strong ability presented by GANs to capture descriptive features \cite{radford2015unsupervised,donahue2016adversarial}. Inspired by that, \cite{zieba2017training} proposes an alternative method for a triplet metric training, called Triplet GAN, based on adapting GANs to perform not only feature extraction but also metric learning.

The main idea behind this approach is to repurpose the discriminator $D$ from classification to a distance learning task, which results in good feature representations during the unsupervised, generative part and supervised, discriminative part of the training process.

To incorporate triplet training into GAN framework, authors propose a modified version of a loss function for a model, specified in Equation \ref{eq:triplet_gan_loss}, with $L(d_+, d_-)$ given as in Equation \ref{eq:triplet_loss}.

\begin{equation}
    L = -V(D,G)-\mathbb{E}_{\mathbf{x}_{q},\mathbf{x}_{+},\mathbf{x}_{-} \sim p_{data}(\mathbf{x}_{q},\mathbf{x}_{+},\mathbf{x}_{-} )}{[\log{(L(d_+, d_-)}]}
    \label{eq:triplet_gan_loss}
\end{equation}

This approach allows improving results in metric learning tasks by allowing to use not only a labeled part of the dataset but by also learn general information about the structure of the data with unsupervised learning on an unlabeled portion of the dataset.

However, learning metric on the discriminative module of the GAN comes with limitations, of which the main one is an inability to perform sampling from the generated representation. Models presented in \cite{donahue2016adversarial,dumoulin2016adversarially} present an extension to the GAN framework, by adding a module, that performs inference on a given data to a latent space representation. A natural question arises: are GANs able to perform latent space embedding that is both regularized by its ability to reconstruct the input and by metric learning approach?


Based on previously presented BiGAN\cite{donahue2016adversarial,dumoulin2016adversarially} model the authors in \cite{zamorski2018semisupervised} address this issue with a presentation of Triplet BiGAN. It combines approaches of BiGAN and Triplet Network \cite{hoffer2015deep} with a joint training objective for Encoder $E$ 
that is trained with both BiGAN and triplet loss. This allows the model to not only learn features from data, but also regularize them with two constraints: the hidden layer encoding tend to be normally distributed (to match the distribution passed to Generator $G$), and embedding of close samples are close to each other in latent space. Representation learned by the Encoder $E$ can be further used in tasks such as retrieval and classification. Triplet BiGAN model is trained in a semi-supervised manner, although it needs as little as 16 labeled samples per class.

Other works worth mentioning are a) on training efficient binary feature representation - \textit{Binary GAN} (BGAN) \cite{song2017binary}, \textit{Binary Regularization Entropy GAN} (BRE-GAN) \cite{Cao2018ImprovingGT}, \textit{Binary GAN} (BinGAN) \cite{zieba2018bingan}, b) on domain adaptation \textit{ARDA} \cite{shen2017adversarial}, c) on learning representation for 3D pointclouds - \textit{3-D GAN} \cite{wu2016learning}, \textit{l-WGAN} \cite{achlioptas2018learning} and \textit{Point Cloud GAN} \cite{li2018point}.

\section{Regularized learning of the discriminator}
In the original paper \cite{goodfellow2014generative} the authors proposed training objective for GANs expressed as a min-max game (Equation \ref{eq:value-fun-G-D-base}). It has been shown, that this approach resulted in highly unstable training \cite{goodfellow2014generative} and authors recommended using an alternative objective instead (Equation \ref{eq:value-fun-G-D-final}). However, even with the modified version, the Generator training often led to vanishing gradients once some of the generated samples were good enough to fool the Discriminator every time, resulting in mode collapse of the Generator. It may be caused by the improper definition of training objective\cite{arjovsky2017wasserstein,berthelot2017began,mao2017least}, where the Generator is rewarded for creating samples indistinguishable from the ones in the training set and not for trying to match the whole distribution of the data.

In \cite{arjovsky2017wasserstein} authors propose a new framework for training GANs called \textit{Wasserstein GAN}. The main improvement over the original framework comes from applying different loss function for the Generator called Wasserstein or Earth-mover distance (Equation \ref{eq:wasserstein_distance}).

\begin{equation}
    \mathrm{EMD}(P_r, P_\theta) = \inf_{\gamma \in \Pi} \ \mathbb{E}_{(x,y) \sim \gamma} \Vert x - y \Vert,
    \label{eq:wasserstein_distance}
\end{equation}
 
 where $\gamma$ is a joined probability distribution between the one modeled by the Generator ($P_\theta$) and the real data distribution $P_r$ and $\Pi$ is the set of all such distributions. The Discriminator's (in context of Wasserstein GAN called as the Critic) role in this scenario is to output a scalar of how real the generated image is, rather than a probability. In practice, the sigmoid activation usually put at the end of the Discriminator (Critic) model is in this scenario omitted.
 
 As the version of the loss presented in Equation \ref{eq:wasserstein_distance} is intractable and thus, impossible to use in this scenario, another formulation, using Kantorovich-Rubinstein duality \cite{rachev1990duality,arjovsky2017wasserstein} is used, as specified in Equation \ref{eq:wasserstein_duality}.
 
 \begin{equation}
    \mathrm{W}(P_r, P_\theta) = \sup_{\lVert f \rVert_L \le 1} \mathbb{E}_{x \sim P_r} [f(x)] - \mathbb{E}_{x \sim P_\theta} [f(x)],
    \label{eq:wasserstein_duality}
\end{equation}
 
 In order for this approach to be effective the function that the Generator G optimizes, must be the n-Lipschitz function\cite{hanin1992kantorovich}, for $n = 1$, i.e. fulfill the constraint given by the Equation \ref{eq:lipschitz_constraint}.
 \begin{equation}
    \frac{|G(x_1) - G(x_2)|}{|x_1 - x_2|} \le 1,
    \label{eq:lipschitz_constraint}
\end{equation}

To satisfy this constraint, the authors of \cite{arjovsky2017wasserstein} suggest clipping weights of the Generator model to the range $[-c, c]$ with the suggested value of the hyperparameter $c=0.01$. However, this method often results in weights distributed near the border values of the range. In \cite{gulrajani2017improved} authors present a new method for satisfying Lipschitz condition, called $gradient penalty$. This method, instead of applying clipping, penalizes the model if the Discriminator (Critic) gradient norm moves away from its target norm value 1.

The Wasserstein GAN framework assumes one iteration update for the Generator weights for five updates of the Critic weights as a way to maintain the stability of the training procedure, but the ratio can be application-specific\cite{arora2017generalization}. The \textit{Boundary Equilibrium GAN}\cite{berthelot2017began} method introduces a procedure to balance the training by the way of maintaining the equilibrium $\mathbb{E}[\mathcal{L}[(G(z))] = \gamma \mathbb{E}[\mathcal{L}(x)]$. It is achieved by the additional parameter $k$ that scales the losses of the Generator $\mathcal{L}_G$ and the Discriminator $\mathcal{L}_D$ as shown in the Equation \ref{eq:began_objective}.

 \begin{equation}
    \mathcal{L}_D = \mathcal{L}(x) - k_t \mathcal{L}(G(z_D))
    \label{eq:began_objective}
\end{equation}
 \begin{equation*}
    \mathcal{L}_G = \mathcal{L}(G(z_G))
\end{equation*}
\begin{equation*}
    k_{t+1} = k_t + \lambda_k (\gamma \mathcal{L}(x) - \mathcal{L}(G(z_G))),
\end{equation*}
where $k$ is a proportion between the Generator and the Discriminator loss at iteration $t$ (with $k_0 = 0$), $\lambda_k$ is the proportional gain for $k$.

Other approaches regularizing training procedure of GAN worth mentioning are \textit{Least Squares GANs} \cite{mao2017least} (applying least squares difference between discriminator loss and the expected outcome), \textit{Spectral Normalization GANs} \cite{miyato2018spectral} (constraining spectral norm of each layer's weights) and \textit{Regularized GANs} \cite{roth2017stabilizing} (adding noise as a regularizer).
\section{Conclusion} 
In this work, we present recent developments in Generative Adversarial Networks research. We explore several selected fields of current research, focusing on the most important milestones, notably in the fields of semi-supervised learning, unsupervised style translation, and representation learning.

\bibliographystyle{splncs04}  
\bibliography{biblography}
\end{document}